# A Large-Scale Car Parts (LSCP) Dataset for Lightweight Fine-Grained Detection

Jie Wang*, Yilin Zhong, Qianqian Cao

BYD Auto Industry Co., LTD.
Photonicsjay@163.com

**Abstract.**
Automotive related datasets have previously been used for training autonomous driving systems or vehicle classification tasks. However, there is a lack of datasets in the field of automotive AI for car parts detection, and most available datasets are limited in size and scope, struggling to cover diverse scenarios. To address this gap, this paper presents a large-scale and fine-grained automotive dataset consisting of 84,162 images for detecting 12 different types of car parts. This dataset was collected from natural cameras and online websites which covers various car brands, scenarios, and shooting angles. To alleviate the burden of manual annotation, we propose a novel semi-supervised auto-labeling method that leverages state-of-the-art pre-trained detectors. Moreover, we study the limitations of the Grounding DINO approach for zero-shot labeling. Finally, we evaluate the effectiveness of our proposed dataset through fine-grained car parts detection by training several lightweight YOLO-series detectors.

## 1 Introduction

Cars play an important role in people's life, providing not only a convenient way of transportation but also bringing economic, cultural, social, and safety benefits [1][2] below. Making cars safer, more intelligent becomes important to car manufacturers. Over the last few years, there has been a significant increase in AI applications in cars. Research in autonomous cars is a rapidly growing field, involving complex software and hardware systems that process vast amounts of data from sensors such as cameras, radars, and GPS [3]. Deep learning algorithms process the collected data and make decisions about steering, braking, accelerating and other actions [4]. To promote driving experience, intelligent voice assistant for vehicles is designed to help drivers control various in-car functions, such as navigation, entertainment, and vehicle settings, with their voice [5]. After-sales car service has also been revolutionized by introducing deep learning predictive analysis. AI-powered predictive maintenance systems can detect potential faults in a vehicle before it breaks down. AI-enabled chat-bots provide an immediate response $24 \times 7$ to customers, making after-sales service more convenient and efficient [6] .

As modern cars become increasingly complex, the ability to quickly and accurately identify and detect car parts has become critical. Car parts detection can be applied in aesthetic design. Well-design car parts can help manufacturers attract more potential customers, increasing sales revenue and gaining market share [7][8]. Besides, vehicle components detecting can ensure the correct use of parts models, accurate assembly positions, and integrity of the parts [9]. In addition, automotive parts detection can also assist in quality control, improving production efficiency and product quality. Moreover, automotive parts detection can be used for component competitive analysis [10]. By detecting and analyzing the components of different vehicles, it is possible to identify part models, materials, specifications, etc., which is of significant importance in developing market strategies and improving product competitiveness [11][12].

Car parts detection research also relies on a large amount of high-quality images, which limits the advance in automobile AI application. To leverage the limitation, unsupervised learning has been carried out. Based on contrastive learning, Segment Anything Model (SAM) [13] is a pre-trained image segmentation model open-sourced by Meta, which has made new breakthroughs in the field of computer vision. SAM can segment novel objects without requiring manual annotations. On the other hand, Grounding DINO [14] has been proposed in the field of object detection. It combines DINO (a unsupervised representation learning method for vision recognition based on Transformers) and visual language alignment methods between images and text. The goal of this method is to use visual-language alignment to enable the model learn the correlation between categories and improve the model's generalization ability across categories and datasets. This method has been verified on the COCO object detection dataset [15] and achieved good results. Although these methods have potential in certain scenarios, its performance still need to be further studied in more extensive datasets and applications. However, these foundation models need high computational resources and storage space, which may be more suitable as offline tools or cloud services to provide prediction results or annotation functions. In this paper, we apply Grounding DINO [14] for automotive annotations and then evaluate its performance on the LSCP dataset.

In order to address the gap, we have provided the LSCP dataset, which consists of 84162 images with 12 types of car parts. The LSCP dataset contains images including both close-up shots of specific details and global views. To validate the usefulness of the LSCP and to encourage the community to explore more for novel research topics, we demonstrate several lightweight YOLO detection experiments with this dataset.

## 2    Related Works

Recently, the availability of large-scale datasets has enabled significant advancements in computer vision and machine learning related to automobiles. Autonomous driving, especially, has made significant progress in recent years, with many self-driving cars testing on public roads. As a real-world driving scenarios dataset, KITTI includes a large number of high-resolution images, LIDAR point clouds, and camera calibration data with annotations, which makes it the most popular autonomous driving dataset [16]. With the rising importance of semantic task in autonomous driving, semantic KITTI offers higher resolution, richer information, and more accurate annotations. A3D dataset [17], also known as ApolloScape 3D dataset, is a large-scale autonomous driving dataset prepared by Baidu's autonomous driving research team, Apollo.

Apart from autonomous driving related datasets, there also exist datasets focusing on fine-grained attributes of vehicles, such as their appearance, brands, and product years. The Stanford Cars Dataset [18] is one of the most well-known related datasets, providing over 16,000 images of cars with classification labels. This dataset has been applied to develop algorithms for car recognition and classification. VehicleX [19] is a large-scale synthetic dataset created by Unity, including 1362 vehicles of various 3D models with fully editable attributes. Vehicle ID captured during the daytime by multiple real-world surveillance cameras distributed in a small city in China contains 26,267 vehicles (221,763 images in total). Each image is attached with an id label corresponding to its identity in real world. Some dataset present attributes for car claims. Comp Cars [20], which was proposed in 2015, has built a large-scale car dataset, including different angle of cars and car parts.

Different from other existing dataset, our dataset has special attributes in car parts dataset. First, our dataset consists of advanced car models with various photograph angles and close-up shots, as well as 12 different parts, making it a large dataset. Second, we use a semi-supervised annotation method to annotate the data. Third, we compare the performance of semi-supervised annotation with manual annotation to further explore the application of large models in automatic labeling.

## 3    LSCP Dataset

The LSCP dataset presented here contains 84162 images of cars from different views, brands and scenarios. This dataset was prepared by combining multiple sources and scraping from public websites to achieve the most complete and accurate dataset as possible. The purpose of this dataset is to offer a valuable benchmark for researchers, automotive professionals, and car enthusiasts who are interested in applying car parts detection for repair, replacement or upgrade purposes.

As shown in Fig. 1 and Fig. 2, the images of car parts are categorized into exterior and interior ones. The exterior one includes wheels, headlight, rear-view mirrors and license, while the interior one contains steering wheels, start buttons, column shifts, combination switchs, inner mirrors, shift knobs and inner handles. We aim to provide more fine-grained types of car parts. Table 1 shows the collection number of different car parts. The LSCP dataset can provide more detailed and accurate information since it can cover more features, and also divide parts into more subcategories. This is valuable for industry professionals, scholars, and researchers as they can gain deeper insights into the characteristics, functions, and applications of car parts.

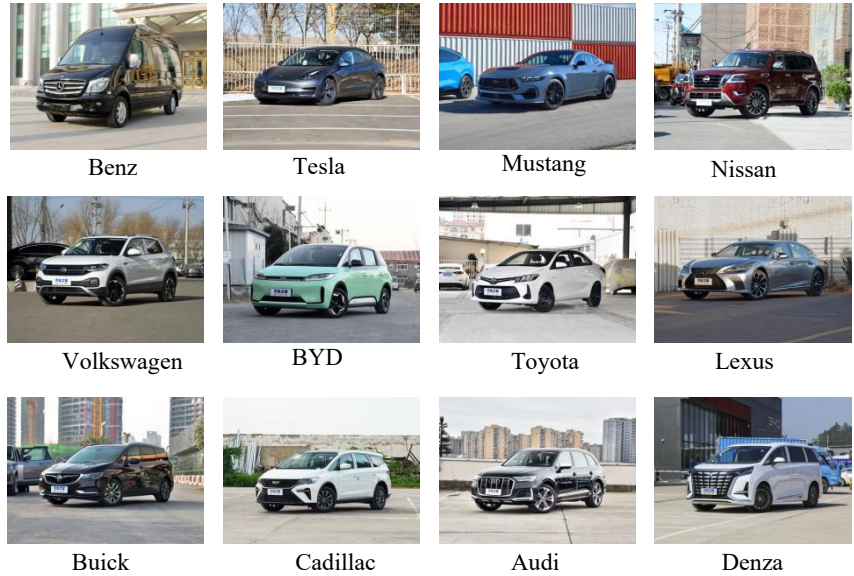

**Fig. 1**. LSCP provides vehicle images with different brands.

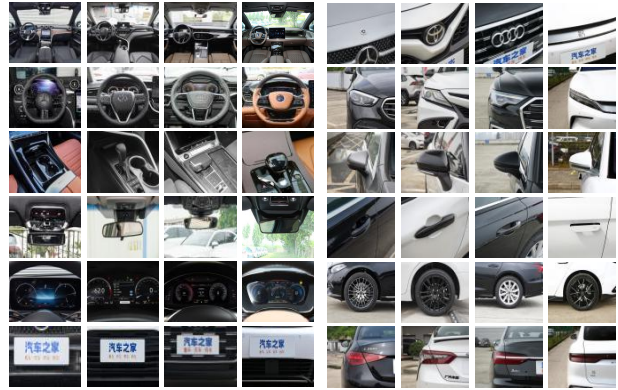

**Fig. 2**. 12 types of car parts in LSCP dataset.

Our LSCP dataset contains images captured from different views, providing more comprehensive and accurate information. Each car in the dataset covers various view-angles including front, black, side, top, bottom, and close-up views. These multiple view-angles can provide more information and enable vision models to better learn the shape, structure and dynamics changes of cars.

**Table 1.** The amount of different car parts.

| Part | No. in total |
|---|---|
| logo | 38647 |
| headlight | 37234 |
| rear-view mirror | 37864 |
| door handle | 39475 |
| wheel | 38942 |
| rear-light | 37468 |
| console | 39143 |
| steering | 36784 |
| gear lever | 38471 |
| inner mirror | 37642 |
| dashboard | 38456 |
| license | 38647 |

## 4    A Review of Object Detection

Object detection involves both localization and classification of object instances by providing class labels and bounding box coordinates. Object localization refers to the process of determining the position and size of any object within an image, which is typically realized using a fully enclosing bounding box determined by a regression model. The IOU value is calculated as the ratio of the intersection area between the predicted and true bounding boxes to the area of their union. Specifically, the formula for calculating the IOU between two bounding boxes A and B is:

$$IOU(A, B) = Area\ of\ Overlap(A, B)\ /\ Area\ of\ Union(A, B) \qquad (1)$$

Object classification involves determining what type of objects in the image. And the output of the classifier is a probability distribution over a predefined set of object classes. Classification loss is commonly defined as cross entropy loss:

$$L_{CE} = -\sum_{i=1}^{n} t_i log(p_i) \qquad (2)$$

As shown in the above formula, $t_i$ represents the probability of the true label being $i$, and $q_i$ represents the probability of the predicted label being $i$ [21].

## 4.1    Lightweight  Object Detector

To facilitate deployment on edge devices with limited computation resources, many lightweight object detectors have been proposed to address this issue, which has a smaller model size and requires fewer computation resources than conventional ones, while maintaining a reasonable level of detection accuracy.

There are various approaches to realize lightweight object detectors, such as using small model architectures, pruning redundant model parameters, or applying efficient training strategies. There exists several popular lightweight object detectors, such as YOLOv5 [22], YOLOx [23], YOLOv7 [24], YOLOv8 [25], NanoDet [26], RTMDet [27]. These models have demonstrated good performance on various object detection datasets while being computationally efficient and suitable for edge devices.

## 4.2    Grounding DINO

Open-world detection aims to accurately detect both known and unkown objects in real scenarios to enhance the perception system's practicality and robustness. This is particularly useful when dealing with continuously evolving environments where new objects may emerge over time. The key to achieving open-set object detection is to fuse language information into the general feature representation of the target objects. For example, GLIP [28] establishes a connection between object detection and textual phrases through contrastive learning, and has shown good performance on both close-set and open-set datasets. Trough combining DINO [29] with CLIP [30], Grounding DINO [14] can detect specified objects based on the textual descriptions. Given the prompts (that is, labels), Grounding DINO [14] can locate and recognize the object, which is a useful tools for annotations.

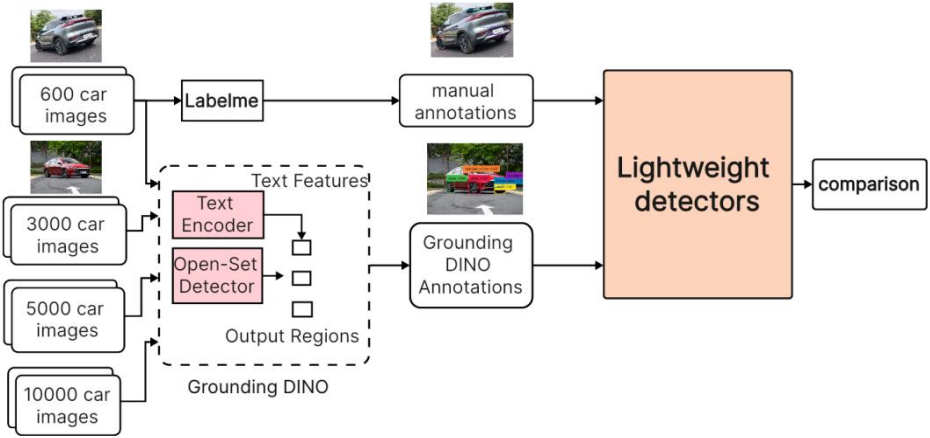

Fig. 3. The procedure of our semi-supervised auto-labeling experiment.

Our experiment flowchart is shown in Fig. 3. Firstly, 600, 3000, 5000, and 10000 images were randomly sampled from the LSCP dataset. Among them, 600 images were manually annotated while all others were auto-labeled using Grounding DINO [14]. Finally, these data were used to train lightweight detectors and the performance of all detectors was compared.

## 5        Semi-Supervised Annotation

This chapter will introduce the data collection process and annotation method of our dataset. We apply a semi-supervised learning method with automatic labeling, which helps to alleviate the tediousness of manual annotation in large datasets.

To evaluate the effectiveness of pre-trained foundation models such as Grounding DINO[14] in object detection annotation, we first manually annotate 600 images and selected 100 of them as a test set. We then input these 100 images along with their corresponding text prompts ("headlights, rear-view mirror, wheel, handle, rear-light, inner mirror, steering wheel, console, gears, dashboard, logo, license") to generate annotations. We evaluate the fine-tuned model using the 100 test images and show the annotation results of Grounding DINO [14] with the manual labels.

**Table 2.** mAP of Grounding DINO[14] as annotators.

| Detector | mAP@50 | mAP@75 |
|----------|--------|--------|
| Grounding DINO | 0.23 | 0.14 |

In this experiment, we also identify the following labeling issues with Grounding DINO [14]. The dependence of Grounding DINO's labeling performance on the text prompts may require significant manual efforts. Moreover, different categories in the text prompts can potentially interfere with each other during the zero-shot labeling. Grounding DINO's annotation performance tends to be more accurate when the text prompts contain fewer distinct classes. As shown in Fig 4, the predicted confidence of object bounding boxes generated by Grounding DINO [14] is not high enough, which may result in object detection misses.

## Experiment and Results

### 5.1        Lightweight Detectors Using Different Labeled Datasets

**Experiment setup.** We sample 3000, 5000, 10000 images from the LSCP dataset to build three training subsets of different sizes for benchmark analysis. Additionally, we randomly select 1000 images from the LSCP dataset to form a validation set. We fine-tune YOLOv8-nano [25] on each of these subsets for 50 epochs with a batch-size of 64. In particular, we also fine-tune the YOLOv8-nano on the fully-manually labeled dataset, the auto-labeled dataset and the above combined one.

For data augmentation, we use random mosaic, rotation, and affine transformations to increase the diversity of the training data. Mosaic is a technique that combines four images into one, enabling object detection in different spatial arrangements. Rotation and affine transformations help the model to handle variations in objects' orientation and scale. We evaluate all the trained models on the validation set using mean average precision(mAP) as the evaluation metric. Moreover, we analyze the inference time of each model on an NVIDIA RTX A4000 GPU.

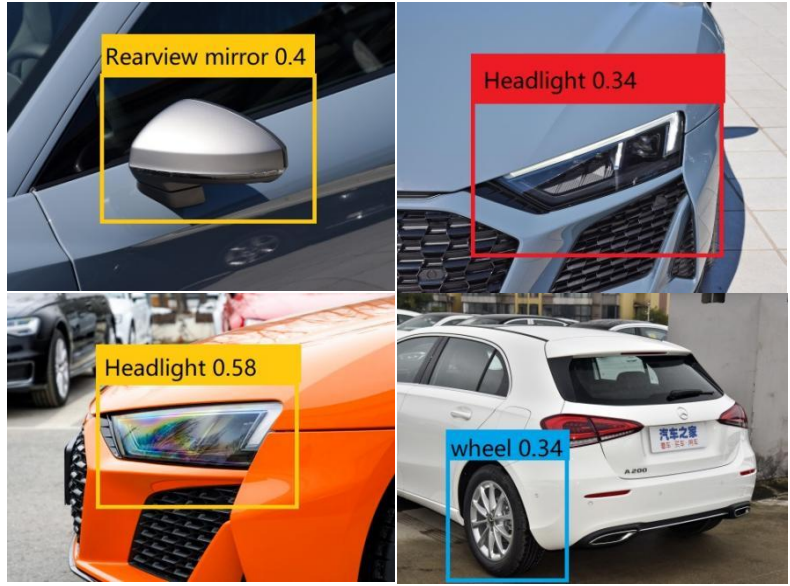

**Fig. 4.** Detection of Grounding DINO [14].

**Results**. As shown in the Table 3, the mAP values of the YOLOv8 detectors increases with the training dataset size, indicating that Grounding DINO based labeling method is effective. Accurate and consistent annotation of object classes, bounding boxes, and other attributes is very crucial for training robust and accurate neural network models. Moreover, it highlights the importance of high-quality and well-annotated datasets for developing and evaluating machine learning algorithms.

**Table 3.** Performance of YOLOv8 trained with datasets of difference sizes.

| Dataset Size | mAP@50 | mAP@75 | mAP_s (small object) | mAP_medium (medium object) | mAP_large (large object) |
|---|---|---|---|---|---|
| 3000 | 0.18 | 0.09 | 0.011 | 0.125 | 0.09 |
| 5000 | 0.21 | 0.10 | 0.01 | 0.10 | 0.13 |
| 10000 | **0.41** | 0.18 | 0.09 | 0.18 | 0.21 |

Table 4 shows the performance of YOLOv8 trained on different annotated datasets. We can infer that introducing the semi-supervised auto-labeling datasets can improve model performance efficiently without extra cost.

**Table 4.** Performance of YOLOv8 trained by datasets of different labeling methods.

| Dataset | mAP@50 | mAP@75 | mAP_s | mAP_m | mAP_l |
|---|---|---|---|---|---|
| Manually-labeled (263) | 0.43 | 0.19 | 0.12 | 0.22 | 0.21 |
| Grounding-DINO labeled (10000) | 0.41 | 0.18 | 0.09 | 0.18 | 0.21 |
| Mixed with above both (10263) | **0.45** | 0.19 | 0.11 | 0.24 | 0.23 |

## 5.2 Evaluation of Different Lightweight Detectors

**Experiment Setup.** To further investigate the application scenarios of our dataset, we sample 20,000 images as a new training set. We then conduct a study on 6 lightweight object detectors pre-trained on COCO dataset [15]. Specifically, we select YOLOv8-nano [25], YOLOv5-nano [22], YOLOv7-tiny [23], NanoDet-m [26] and RTMdet [27]. We use a batch-size of 64 and fine-tune the model for 20 epochs, with an initial learning rate of 0.001. We also apply data augmentation techniques such as mosaic, random cropping, random flipping, HSV random augmenting and scaling to the input images during training to increase the diversity. During training, we monitor the loss function and evaluation metrics such as mAP and recall to track the model training. Finally, we fine-tune the model on the smaller validation set before testing on a held-out test set to evaluate the final performance of the model.

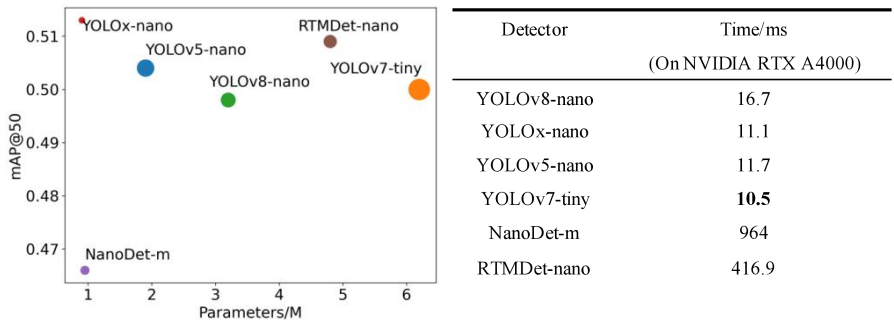

| Detector | Time/ms |
|---|---|
| | (On NVIDIA RTX A4000) |
| YOLOv8-nano | 16.7 |
| YOLOx-nano | 11.1 |
| YOLOv5-nano | 11.7 |
| YOLOv7-tiny | **10.5** |
| NanoDet-m | 964 |
| RTMDet-nano | 416.9 |

**Fig. 5.** Comparison of different lightweight detectors.

**Results.** According to Fig. 5, YOLOx-nano [25] achieves the highest mAP, while YOLOv7-tiny [24] owns the fastest inference speed. Both metrics are the important indicators of a detection model's performance and efficiency. A high mAP indicates that the model is accurate in detecting objects, while a fast inference speed means that the model can process input images quickly, which is especially important for real-time applications. Ultimately, the choice of which model to use will depend on the specific requirements of the application, balancing the trade-off between accuracy and inference speed.

# 6    Conclusion

In conclusion, our paper presents three significant contributions related to automotive detection and annotation. First, we propose a large-scale and fine-grained dataset of car parts which can be used for training and testing automotive detection models. This dataset offers a crucial testbed for automotive researchers and industry professionals who are developing automotive intelligent algorithms. Second, we introduce a novel semi-supervised auto-labeling method which utilizes the pre-trained foundation model to improve the efficiency and accuracy of label annotation. Our experimental results demonstrate the effectiveness of this approach and also analyze the limitations and challenges of such zero-shot learning methods in real-world scenarios. Finally, we provide a benchmark with a series of lightweight detection models on our proposed LSCP dataset. And this benchmark provides valuable insights into the performance and efficiency of YOLO-series models, providing guidelines of model selection for researchers. Overall, our work potentially contributes to advancing the automotive intelligent field in multiple ways: providing a new dataset, proposing a novel auto-labeling method, and evaluating various lightweight detection models on our datasets. Although there are still some challenges and limitations to overcome, we believe that our study will inspire further developments in this field, leading to even more precise and efficient object detection algorithms.